\documentclass[a4paper,11pt]{article}

\usepackage{times}
\usepackage{fullpage}
\usepackage{url}
\usepackage{stmaryrd}
\usepackage{amsmath}
\usepackage{titlesec}

\title{Exploring a type-theoretic approach to \\accessibility constraint
  modelling}
\author{Sylvain Pogodalla\\LORIA/INRIA}
\date{}

\newcommand\inter[1]{\llbracket #1 \, \rrbracket}
\newcommand\interit[1]{\inter{\textit{#1}}}
\newcommand\arrow{\rightarrow}
\newcommand\SC{\textit{s}}
\newcommand\VP{\textit{VP}}

\newcommand\sub[1]{\textbf{\scriptsize #1}}
\newcommand\nil{\texttt{nil}}
\newcommand\coord{._{\textbf{c}}}
\newcommand\subord{._{\textbf{s}}}

\titlespacing{\paragraph}{0pt}{*2}{*1}
\newcommand\mysection[1]{\paragraph{#1}}

\begin{document}



\maketitle


\mysection{Accessibility constraints}

When dealing with anaphora resolution, anaphoric expressions are often
considered to have access to a restricted set of discourse referents.
The choice of the relevant discourse referent inside this set may
depends on various kinds of information (morphosyntactic features,
semantic features, salience\ldots). But theories that give a dynamic
interpretation of language also provide some ways to restrict the set
of discourse referents by stating accessibility constraints.

The description of these constraints depends on the chosen formal
theory of discourse interpretation. For instance, DRT~\cite{kamp93}
describes the accessibility constraints using DRS subordination.
Taking into account the hierarchical structure of discourse,
SDRT~\cite{asher2003} uses DRS subordination together with an outscoping
relation between discourse units. In both cases, the constraints are
based on the \emph{structure} of the discourse: for DRT, the structure
arises from the negation while for SDRT it also arises from the nature
of the relation between discourse units. In the latter case, it gives
a formal definition of the Right Frontier Constraint
(RFC)~\cite{asher08short}.

\cite{degroote06} proposes a modelling of DRT that fits Montague's
semantic framework. In this approach, the logical quantifiers have
their standard scoping definitions and free and bound variables also
have their standard definitions. One of the advantages is to use
standard notions (such as De~Bruijn's indices for $\lambda$-calculus)
for implementing variable renaming in semantic systems based on
$\lambda$-calculus, instead of implementing specific and complex
work-around. \cite{degroote06} also advocates its independence from
any specific theory. While~\cite{degroote06} exemplifies the DRT view
on accessibility constraints, the work we propose here consists in
exploring this approach in modelling two other kinds of accessibility
constraints: the accessibility to proper nouns in DRT and the RFC.

\mysection{A Montagovian flavour of DRT} In~\cite{degroote06}, the
syntactic type of sentences $\SC$ is interpreted with the semantic
type: $\inter{\SC}=\gamma \arrow (\gamma \arrow t) \arrow t$.  It
means that (the semantic interpretation of) a sentence, instead of
being a proposition, requires two arguments to produce a proposition.
The first argument is its left context (we can think about it as the
current set of discourse referents). The second argument is its right
context, or its \emph{continuation}: something able to return a
proposition if fed with a (potentially updated) set of discourse
referents.

Two sentences $s_1$ and $s_2$ are combined in the following way:
$\inter{s_1. s_2}=\lambda e \phi.\inter{s_1} \, e \, (\lambda
e'.\inter{s_2} \, e' \, \phi)$. It means that the result of the
combination of $s_1$ and $s_2$ takes as input an environment $e$ and a
continuation $\phi$. This environment is the same as the one $s_1$ has
access to. So $\inter{s_1}$ takes $e$ as first parameter. Then, the
continuation of $s_1$ is made of something taking $e'$ as left
environment\footnote{Note that the actual value of $e'$ will be
  provided by the semantic recipe of $s_1$. That is $s_1$ can choose
  to update or not $e$ before it gives it as parameter to its
  continuation.} and the final result is the one we get by feeding
$s_2$ with the new left environment $e'$ and the same continuation as
the $s_1 .  s_2$ combination.

With this interpretation and the right semantic recipes
(see~\cite{degroote06} for the detailed lexical semantics) we get the
following interpretation for \textit{John loves a woman}: $\lambda e
\phi.\exists y. \textbf{woman y} \land \textbf{love} \, \textbf{j} \,
y \land \phi \, (y::e)$ where $::$ (of type $e \arrow \gamma \arrow
\gamma$) is a function that adds a discourse referent to a set of
discourse referents. Note that the continuation of this sentence,
$\phi$, will be provided with the $y$ discourse referent, bound in the
standard way by the existential quantifier.


\mysection{Modelling other accessibility constraints}

To state as in DRT that negation blocks the access to discourse
referents,~\cite{degroote07} interprets: $\inter{(\textit{doesn't
    \VP})S}=\lambda e \phi. \neg ((\VP \, S)\,e\, (\lambda e'.\top))
\land \phi \, e$. The continuation of the sentence ($\phi$) has no
access to what $\VP$ and $S$ could introduce as discourse referents,
it only has access to the discourse referents of $e$\footnote{$\top$
  stands for true. That is, $\lambda e.\top$ is the continuation that
  always returns true.}.

Note that an alternative could have been: $\interit{(\text{doesn't
    \VP})S}=\lambda e \phi. \neg ((\VP \, S)\,e\, (\lambda e'. \phi \,
e'))$.  But in this case, the continuation would also have been in the
scope of the negation: the rest of the discourse would also be
negated, which is not what we expect.

We propose to introduce the type of logical connectives $\kappa=t
\arrow t \arrow t$ and to interpret the syntactic type of sentences as
$\inter{\SC}=\kappa \arrow \gamma \arrow (\kappa \arrow \gamma \arrow
t) \arrow t$. Then negation can be interpreted as: $\inter{
  (\textit{doesn't}\,\VP)\,S}= \lambda c e \phi. \neg ((\VP\,S)\, (\neg
c)\, e (\lambda c' e'. \neg (\phi\,c'\,e)))$. Without all the details,
if we think as $c$ being the logical connective $\land$, we have the
result to be interpreted as $\neg ((\VP\,S) \lor \neg (\phi \,
e))\equiv (\neg (\VP\,S))\land (\phi\,e)$ which is now the expected
result. Interestingly, we see that it could also give access to the
discourse referents introduced by $\VP \, S$ to $\phi$ using the
continuation $\lambda c' e'. \neg (\phi \, c' \, e')$. This, of
course, is not expected, except for proper nouns (which are ``always
on top'' in DRT). We then see how to add an additional left
environment parameter for proper nouns (say $e_1$) so that negation
blocks the discourse referents introduced by existentials (in $e_2$)
but propagates the ones introduced by proper nouns. With the new value
of $\inter{\SC} = \kappa \arrow \gamma \arrow \gamma \arrow (\kappa
\arrow \gamma \arrow \gamma \arrow t) \arrow t $ and of sentence
composition: $\inter{s_1.s_2} = \lambda c e_1 e_2
\phi.\inter{s_1}\,c\,e_1\,e_2(\lambda c' e'_1 e'_2.\inter{s_2} c' e'_1
e'_2 \phi)$)\footnote{We use here a currified notation that introduces
  as many arrows as functions have parameters. We could of course use
  a product or record type to keep only one parameter.}, the negation
is now interpreted as: $\interit{doesn't} = \lambda V S c e_1 e_2 \phi.
\neg ((V\,S)\, (\neg c)\, e_1 \, e_2(\lambda c' e'_1 e'_2.  \neg (\phi
\, c' \, e'_1 \, e_2)))$.  Together with the following
interpretations:$$\begin{array}{ll} \interit{John} & = \lambda P c e_1
  e_2
  \phi. P\,\textbf{j}\, c\,(\textbf{j}::e_1) \,e_2\,\phi\\
  \interit{own} & = \lambda O S .S(\lambda x. O(\lambda y c' e'_1 e'_2
  \phi'. c' (\textbf{own}\,x\,y)(\phi'\,c'\,e'_1\,e'_2)))\\
  \interit{car} & = \lambda x c e_1 e_2 \phi. c(\textbf{car}\, x)(\phi\,c\,e_1\,e_2)\\
  \interit{a} & = \lambda P Q c e_1 e_2 \phi. \exists x. [\lambda
  \phi'.(P\,x \, c \, e_1 \,e_2 \,\phi') \land (Q
  \, x \, c \, e_1 \, e_2 \, \phi')] (\lambda c' e'_1 e'_2 . \phi c e'_1 (x::e'_2))\\
  \interit{it} & = \lambda P .
  P\,(\texttt{sel}\,(e_1 \cup e_2))\\
  \interit{is} & = \lambda A S .S(\lambda x c' e'_1 e'_2 \phi'. c' (A
  (\lambda y
  c'' e''_1 e''_2 \phi''.\top) x \,c'\,e'_1 \,e'_2 \, \phi') (\phi'\,c'\,e'_1\,e'_2))\\
  \interit{red} & = \lambda P x c e_1 e_2 \phi. (P\,x \, c \, e_1
  \,e_2 \,\phi) \land (\textbf{red}\,x)
\end{array}
$$
we can interpret:$$
\begin{array}{ll}
  \interit{own} (\interit{a}\interit{car}) & = \lambda S.S(\lambda x c e_1 e_2 \phi.\exists y.[\lambda \phi'. (c(\textbf{car}\,y)(\phi'\,c\,e_1\,e_2))  \land (c (\textbf{own}\,x\,y)(\phi' c e_1 e_2))]\\
  & \qquad \quad  (\lambda c' e'_1 e'_2.\phi\,c\,e'_1(y::e'_2)))\\
  & \equiv \lambda S.S(\lambda x c e_1 e_2 \phi.\exists y.[\lambda \phi'. c((\textbf{car}\,y) \land (\textbf{own}\,x\,y))(\phi'\,c\,e_1\,e_2)] \\
  & \qquad \quad  (\lambda c' e'_1 e'_2.\phi\,c\,e'_1(y::e'_2)))\\
  & = \lambda S.S(\lambda x c e_1 e_2 \phi.\exists y.c((\textbf{car}\,y) \land (\textbf{own}\,x\,y))(\phi\,c\,e_1\,(y::e_2))) \\
\end{array}
$$
thanks to the following equivalence ($c$ is either $\land$ or $\lor$):
$$(c(\textbf{car}\,y)(\phi'\,c\,e_1\,e_2)) \land (c
(\textbf{own}\,x\,y)(\phi' c e_1 e_2))) \equiv c (\textbf{car}\,y
\land \textbf{own}\,x\,y )(\phi' c e_1 e_2)$$ Finally, a discourse $d$
such as \textit{John doesn't own a car. It is red} with the empty
context \texttt{nil} and the empty continuation $\phi_e=\lambda c e_1
e_2.\neg (c\, \top \bot)$ (which returns $\top$ under the $\land$
connective and $\bot$ under the $\lor$ connective) is interpreted as:
$$
\begin{array}{ll}
  \interit{d}\,(\land)\,\nil\,\nil\,\phi_e & =
  \neg (\exists y. (\textbf{car}\,y \land \textbf{own} \, \textbf{j}\,y)  \lor (\neg (\textbf{red}(\texttt{sel}((\textbf{j}::\nil)\cup\nil))) \lor \bot)) \\
  & =
  \neg (\exists y. (\textbf{car}\,y \land \textbf{own} \, \textbf{j}\,y) \lor (\neg (\textbf{red}(\texttt{sel}(\textbf{j}::\nil))))) \\
  & \equiv
  (\neg \exists y. (\textbf{car}\,y \land \textbf{own} \, \textbf{j}\,y)) \land \textbf{red}(\texttt{sel}(\textbf{j}::\nil))
\end{array}
$$
It shows that whereas the discourse referent introduced by $y$ is not
accessible to the $\texttt{sel}$ operator, the discourse referent
introduced by $\textbf{j}$ would be accessible, in spite of the
negation.

We extend this approach to model the RFC, introducing a parameter for
the type of discourse relations: subordinating or coordinating, and
two ways of combining sentences: $s_1 ._\sub{c} s_2 $ and $s_1
._\sub{s} s_2 $ instead of just one $s_1 . s_2$. The interpretation of
these two combinations manage the set of discourse referents so that
the environment given to $s_2$ only includes the set of discourse
referents defined by the RFC. As for now, we only deal with discourse
structures that can be described by a formula (or syntactic tree) made
of the $._\sub{c}$ and of the $._\sub{s}$ connectives.  Discourse pops
or discourse structure that are not trees, for instance, are not yet
considered.

We consider a new type $\kappa = \gamma \arrow \gamma \arrow \gamma$
and $\inter{\SC} = \kappa \arrow \gamma \arrow \gamma \arrow (\kappa
\arrow \gamma \arrow \gamma \arrow t) \arrow t$, and we define
$\textbf{Coord}= \lambda e_1 e_2.e_2$ and $\textbf{Sub} = \lambda e_1
e_2.e_1 \cup e_2$. We give the following definition of sentence
composition:
$$
\begin{array}{ll}
  \inter{s_1 \subord s_2} & = \lambda c e_1 e_2 \phi. \inter{s_1}\, c\,
  e_1\,e_2 (\lambda c' e'_1 e'_2.\inter{s_2}\,
  \textbf{Sub}\,e'_1\,(c\,e_1\, e_2)\,\phi)\\
  \inter{s_1 \coord s_2} & = \lambda c e_1 e_2 \phi. \inter{s_1}\, c\,
  e_1\,e_2 (\lambda c' e'_1 e'_2.\inter{s_2}\,
  \textbf{Coord}\,e'_1\,(c\,e_1\, e_2)\,\phi)
\end{array}$$
Then
$$
\begin{array}{ll}
  \inter{s_1 \coord (s_2 \coord s_3)} & =   \lambda c e_1 e_2 \phi. \inter{s_1}\,c\,e_1\,e_2 (\lambda c' e'_1 e'_2
  .\inter{s_2}\,\textbf{Coord}\,e'_1\,(c\,e_1\,e_2)\\ & \qquad(\lambda c''' e'''_1 e'''_2.\inter{s_3}\,\textbf{Coord}\,e'''_1\,(c\,e_1\,e_2)\,\phi))
\\
  \inter{s_1 \coord (s_2 \subord s_3)} & =   \lambda c e_1 e_2 \phi. \inter{s_1}\,c\,e_1\,e_2 (\lambda c' e'_1 e'_2
  .\inter{s_2}\,\textbf{Coord}\,e'_1\,(c\,e_1\,e_2) \\ & \qquad(\lambda c''' e'''_1 e'''_2.\inter{s_3}\,\textbf{Sub}\,e'''_1\,(c\,e_1\,e_2)\,\phi))
\\
  \inter{s_1 \subord (s_2 \coord s_3)} & =   \lambda c e_1 e_2 \phi. \inter{s_1}\,c\,e_1\,e_2 (\lambda c' e'_1 e'_2
  .\inter{s_2}\,\textbf{Sub}\,e'_1\,(c\,e_1\,e_2) \\ & \qquad(\lambda c''' e'''_1 e'''_2.\inter{s_3}\,\textbf{Coord}\,e'''_1\,(e'_1\cup(c\,e_1\,e_2))\,\phi))
\\
  \inter{s_1 \subord (s_2 \subord s_3)} & =   \lambda c e_1 e_2 \phi. \inter{s_1}\,c\,e_1\,e_2 (\lambda c' e'_1 e'_2
  .\inter{s_2}\,\textbf{Sub}\,e'_1\,(c\,e_1\,e_2) \\ & \qquad(\lambda c''' e'''_1 e'''_2.\inter{s_3}\,\textbf{Sub}\,e'''_1\,(e'_1\cup(c\,e_1\,e_2))\,\phi))
\end{array}
$$
which means that in addition to the environment introduced by their
previous discourse unit ($e'_1$ for $s_2$ and $e''_1$ for $s_3$),
$s_2$ and $s_3$ have access to $(c\,e_1\,e_2)$ whose value is: either
$e_2$ if the whole part of the discourse is in a coordinating relation
with what comes before (and then only $s_1$ should access $e_1$), or
$e_1 \cup e_2$ if the whole part is in a subordinating relation with
what comes before (and the previous unit dominates $s_1$, $s_2$ and
$s_3$ and each have access to $e_1$ and $e_2$).

\mysection{Conclusion} 

We show how~\cite{degroote06}'s approach is flexible enough to model
various accessibility constraints, such as the ones for discourse
referents introduced by proper nouns or by the hierarchical structure
of the discourse. Moreover, it proves to be able to combine the
different constraints. We hope this could be helpful to give an
account of the hierarchy of referential expressions and their adequacy
to the RFC~\cite{asher08short} by combining various constraints for
pronouns or definite descriptions for instance.

{\small \bibliographystyle{plain} \bibliography{general,degroote} }

\end{document}